\documentclass[12pt]{spieman}  

\usepackage{amsmath,amsfonts,amssymb}
\usepackage{graphicx}
\usepackage{setspace}
\usepackage{tocloft}

\usepackage{graphicx}
\usepackage{bm}
\usepackage{mathrsfs}
\usepackage{amsmath}
\usepackage{amssymb}
\usepackage{algorithm}
\usepackage{algorithmic}
\usepackage{array}
\usepackage{bm}
\usepackage{booktabs}
\usepackage{caption3}
\usepackage{url}
\usepackage{multirow}
\usepackage{xcolor}
\usepackage{lineno}
\newcommand{\etal}{\textit{et al.~}}

\title{2DR$_1$-PCA and 2DL$_1$-PCA: two variant 2DPCA algorithms based on none L$_2$ norm}

\author[a,b]{Xing Liu}
\author[a,b,*]{Xiao-Jun Wu}
\author[a,b]{Zi-Qi Li}
\affil[a]{Jiangnan University, School of Internet of Things Engineering, Lihu Avenue, Wuxi, China}
\affil[b]{Jiangsu Provincial Laboratory of Pattern Recognition and Computational Intelligence, Lihu Avenue, Wuxi, China}

\cftpagenumbersoff{figure}
\cftpagenumbersoff{table}

\begin{document}


\maketitle

\begin{abstract}
In this paper, two novel methods: 2DR$_1$-PCA and 2DL$_1$-PCA are proposed for face recognition. Compared to the traditional 2DPCA algorithm, 2DR$_1$-PCA and 2DL$_1$-PCA are based on the R$_1$ norm and L$_1$ norm, respectively. The advantage of these proposed methods is they are less sensitive to outliers. These proposed methods are tested on the ORL, YALE and XM2VTS databases and the performance of the related methods is compared experimentally.
\end{abstract}

\keywords{Face recognition; 2DR$_1$-PCA; 2DL$_1$-PCA; R$_1$ norm; L$_1$ norm}

{\noindent \footnotesize\textbf{*}Xiao-Jun Wu,  \linkable{wu\_xiaojun@jiangnan.edu.cn} }

\begin{spacing}{2}   

\section{Introduction}
\label{sec:1}  
Feature extraction by dimensionality reduction is a critical step in pattern recognition. Principal component analysis (PCA) is a classic method for dimensionality reduction in the field of face recognition, which was proposed by Turk and Pentland in Ref.~\citenum{turk1991eigenfaces}. Yang \etal \cite{yang2004two} presented two-dimensional PCA (2DPCA) to improve the efficiency of feature extraction, in which image matrices were used directly. Two-dimensional weighted PCA (2DWPCA) was developed in Ref.~\citenum{nhat2005two} to improve the performance of 2DPCA. The complete 2DPCA method was presented in Ref.~\citenum{xu2006complete} to reduce the feature coefficients needed for face recognition compared to 2DPCA. In kernel PCA (KPCA)~\cite{yang2000face}, samples were mapped into a high dimensional and linearly separable kernel space and then PCA was employed for feature extraction. Chen \etal \cite{chen2017kpca} presented a pattern classification method based on PCA and KPCA (kernel principal component analysis), in which within-class auxiliary training samples were used to improve the performance. Liu \etal \cite{liu2011eca} proposed a 2DECA method, in which features are selected in 2DPCA subspace based on the Renyi entropy contribution instead of cumulative variance contribution. Moreover, some approaches based on linear discriminant analysis (LDA) were explored~\cite{xiao2004new,zheng2006nearest,zheng2006reformative}.

Contrast to the above L$_2$ norm based methods, Kwak~\cite{kwak2008principal} developed L$_1$-PCA by using L$_1$ norm. Ding \etal \cite{ding2006r} proposed a rotational invariant L$_1$ norm PCA (R$_1$-PCA). These none L$_2$ norm based algorithms are less sensitive to the presence of outliers.

In this paper we propose 2DR$_1$-PCA and 2DL$_1$-PCA algorithms for face recognition by utilizing the advantages of L$_1$ norm method and 2DPCA. Instead of using image vectors in R$_1$-PCA and L$_1$-PCA, we use image matrices in 2DR$_1$-PCA and 2DL$_1$-PCA directly for features extraction. Compared to the 1-D methods, the corresponding 2-D methods have two main advantages: higher efficiency and recognition accuracy. We extend R$_1$-PCA and L$_1$-PCA to their two dimensional case and the 2DR$_1$-PCA and 2DL$_1$-PCA methods are proposed.

This paper is organized as follows: We give a brief introduction to the R$_1$-PCA and L$_1$-PCA algorithms in Section 2. In Section 3, the 2DR$_1$-PCA and 2DL$_1$-PCA algorithms are proposed. In Section 4, the mentioned methods are compared through experiments. Finally, conclusions are drawn in Section 5.

\section{Fundamentals of subspace methods based on none L$_2$ norm}
\label{sec:2}
In this paper, we use $X=\left[x_{1}, x_{2}, x_{3}, \ldots, x_{i}, \ldots, x_{n}\right]$ to denote the training set of 1-D methods, where $x_{i}$ is a $d$-dimensional vector.

\subsection{R$_1$-PCA}
\label{sec:2.1}
R$_1$-PCA algorithm tries to find a subspace by minimizing the following error function
\begin{equation}\label{equ:1}
E_{R_{1}}=\|X-W V\|_{R_{1}}
\end{equation}
where $W$ is the projection matrix, $V$ is defined as $V=W^{T} X$, and $\|\cdot\|_{R_{1}}$ denotes the R$_1$ norm, which is defined as
\begin{equation}\label{equ:2}
\|X\|_{R_{1}}=\sum_{i=1}^{n}\left(\sum_{j=1}^{d} x_{j i}^{2}\right)^{\frac{1}{2}}
\end{equation}

In R$_1$-PCA algorithm, the training set $X$ should be centered, i.e.,$x_{i}=x_{i}-\bar{x}$, where $\bar{x}$ is the mean vector of $X$, which is given by $\bar{x}=\frac{1}{n} \sum_{i=1}^{n} x_{i}$.

The principal eigenvectors of the R$_1$-covariance matrix is the solution to R$_1$-PCA algorithm. The weighted version of R$_1$-covariance matrix is defined as
\begin{equation}\label{equ:3}
C_{r}=\sum_{i} \omega_{i} x_{i} x_{i}^{T}, \omega_{i}^{\left(L_{1}\right)}=\frac{1}{\left\|x_{i}-W W^{T} x_{i}\right\|}
\end{equation}

The weight has many forms of definitions. For the Cauchy robust function, the weight is
\begin{equation}\label{equ:4}
\omega_{i}^{(C)}=\left(1+\left\|x_{i}-W W^{T} x_{i}\right\|^{2} / c^{2}\right)^{-1}
\end{equation}

The basic idea of R$_1$-PCA is starting with an initial guess $W^{(0)}$ and then iterate $W$ with the following equations until convergence
\begin{equation}\label{equ:5}
\left\{\begin{array}{c}{W^{\left(t+\frac{1}{2}\right)}=C_{r}\left(W^{(t)}\right) W^{(t)}} \\ {W^{(t+1)}=\text {orthoronalize }\left(W^{\left(t+\frac{1}{2}\right)}\right)}\end{array}\right.
\end{equation}

The concrete algorithm is given in Algorithm~\ref{alg:1}.
\begin{algorithm} 
	\caption{R$_1$-PCA algorithm} 
	\label{alg:1} 
	\begin{algorithmic}[1]
		\REQUIRE The training set $X=\left[x_{1}, x_{2}, x_{3}, \cdots, x_{i}, \cdots, x_{n}\right]$ and the subspace dimension $k$. Then the training set $X$ is centered.
		\STATE Initialization: Compute standard PCA and obtain $W_0$. Set $W=W_0$.
		\STATE Calculate the Cauchy weight:
		
		Compute residue $S_{i}=\sqrt{x_{i}^{T} x_{i}-x_{i}^{T} W W^{T} x_{i}}$
		
		Compute $c=m e \operatorname{dian}\left(s_{i}\right)$
		
		Compute $\omega_{i}=\left(1+\left\|x_{i}-W W^{T} x_{i}\right\|^{2} / c^{2}\right)^{-1}$
		\STATE Calculate the covariance matrix: $C_{r}=\sum_{i} \omega_{i} x_{i} x_{i}^{T}$
		\STATE Update $W$:
		
		$W^{\left(t+\frac{1}{2}\right)}=C_{r}\left(W^{(t)}\right) W^{(t)}$
		
		$W^{(t+1)}=$ orthoronalize $\left(W^{\left(t+\frac{1}{2}\right)}\right)$
		\STATE Convergence check:
		
		If $W^{t+1} \neq W^{t}$, go to Step 4.
		
		Else go to Step 6.
		\STATE Calculate the uncentered data: $\forall i, x_{i}=x_{i}+\bar{x}$.
		\STATE Projection: $V=W^{T} X$.
		\ENSURE $W$ and $V$.
	\end{algorithmic} 
\end{algorithm}

\subsection{L$_1$-PCA}
\label{sec:2.2}
The L$_1$ norm is used in L$_1$-PCA for minimizing the following error function
\begin{equation}\label{equ:6}
E_{L_{1}}=\|X-W V\|_{L_{1}}
\end{equation}
where $W$ is the projection matrix, $V$ is defined as $V=W^{T} X$, and $\|\cdot\|_{L_{1}}$ denotes the L$_1$ norm, which is defined as
\begin{equation}\label{equ:7}
\|X\|_{L_{1}}=\sum_{i=1}^{d} \sum_{j=1}^{n}\left|X_{i j}\right|
\end{equation}

In order to obtain a subspace with the property of robust to outliers and invariant to rotations, the L$_1$ norm is adopted to maximize the following equation
\begin{equation}\label{equ:8}
W^{*}=\max _{W}\left\|W^{T} X\right\|_{L_{1}}, \text { subject to } W^{T} W=I
\end{equation}

It is difficult to solve the multidimensional version. Instead of using projection matrix $W$, a column vector $w$ is used in equation (\ref{equ:8}) and the following equation is obtained
\begin{equation}\label{equ:9}
w^{*}=\max _{w}\left\|w^{T} X\right\|_{L_{1}}, \text { subject to }\|w\|_{2}=1
\end{equation}

Then a greedy search method is used for solving (\ref{equ:9}), which is summarized in Algorithm~\ref{alg:2}.
\begin{algorithm} 
	\caption{L$_1$-PCA} 
	\label{alg:2} 
	\begin{algorithmic}[1]
		\REQUIRE The training set $X=\left[x_{1}, x_{2}, x_{3}, \cdots, x_{i}, \cdots, x_{n}\right]$.
		\STATE Initialization: Initialize $w_0$ by random numbers. Then set $w(0)=w(0) /\|w(0)\|_{2}, t=0$
		\STATE Polarity check: $\forall i \in\{1,2,3, \cdots, n\},$ if $w^{T}(t) x_{i}<0, p_{i}(t)=-1,$ otherwise, $p_{i}(t)=1$.
		\STATE Flipping and maximization: Set $t=t+1, w(t)=\sum_{i=1}^{n} p_{i}(t-1) x_{i}, w(t)=w(t) /\|w(t)\|_{2}$.
		\STATE Convergence check:
		
		If $w(t) \neq w(t-1)$, go to step 2.
		
		Else if $i$ exists such that $w^{T}(t) x_{i}=0$, set $w(t)=(w(t)+\Delta w) /\|w(t)+\Delta w\|_{2}$, where
		$\Delta w$ is a small nonzero random vector. Go to step 2.
		
		Otherwise, set $w^{*}=w(t)$ and stop.
		\ENSURE The projection vector $w$.
	\end{algorithmic} 
\end{algorithm}

One best feature is extracted by the above algorithm. In order to obtain a $k$ dimensional projection matrix instead of a vector, an algorithm based on the greedy search method is given as follows

$w_{0}=\mathbf{0},\left\{x_{i}^{0}=x_{i}\right\}_{i=1}^{n}$

For $j=1$ to $k$

$\quad \forall i \in\{1,2,3 \cdots, n\}, x_{i}^{j}=x_{i}^{j-1}-w_{j-1}\left(w_{j-1}^{T} x_{i}^{j-1}\right)$

$\quad \quad$ Apply the L$_1$-PCA procedure to $X^{j}=\left[x_{1}^{j}, \cdots, x_{n}^{j}\right]$ to find $w_{j}$

End

\section{2DR$_1$-PCA and 2DL$_1$-PCA algorithms}
\label{sec:3}
In 2-D methods, $F=\left[F_{1}, F_{2}, F_{3}, \cdots, F_{i}, \cdots, F_{n}\right]$ is used to denote the training set, where $F_{i}$ is a $r \times n^{'}$ matrix.

\subsection{2DR$_1$-PCA}
\label{sec:3.1}
In this paper we propose 2DR$_1$-PCA algorithm, in which we iterate the projection matrix $W$ with an initial matrix $W^{0}$ until convergence.

First, the training set $F$ is centered, i.e., $F_{i}=F_{i}-\bar{F}$, where $\bar{F}$ is the mean matrix of $F$, defined as $\bar{F}=\frac{1}{n} \sum_{i=1}^{n} F_{i}$.

The R$_1$ covariance matrix is defined as
\begin{equation}\label{equ:10}
C_{r}=\sum_{i=1}^{n} \omega_{i} F_{i} F_{i}^{T}
\end{equation}

The Cauchy weight is defined as
\begin{equation}\label{equ:11}
\left\{\begin{array}{c}
{W_{i}^{\left(C\right)}=\left(1+\left\| F_{i}-WW^{T}F_{i} \right\|_{F}^{2} / c^{2} \right)^{-1}} \\ {c=median\left(s_{i}\right)}
\end{array}\right.
\end{equation}

The residue $s_{i}$ is defined as
\begin{equation}\label{equ:12}
s_{i}=\left\|F_{i}-W W^{T} F_{i}\right\|_{F}
\end{equation}

After obtaining the eigenvectors of $C_{r}$, the iterative formula is similar to which used in the R$_1$-PCA algorithm
\begin{equation}\label{equ:13}
\left\{\begin{array}{c}
{W^{\left(t+\frac{1}{2}\right)}=C_{r}\left(W^{(t)}\right) W^{(t)}} \\ {W^{(t+1)}=orthoronalize\left(W^{\left(t+\frac{1}{2}\right)}\right)}
\end{array}\right.
\end{equation}

The 2DR$_1$-PCA algorithm is outlined in Algorithm~\ref{alg:3}.
\begin{algorithm} 
	\caption{The 2DR$_1$-PCA algorithm} 
	\label{alg:3}
	\begin{algorithmic}[1]
		\REQUIRE The training set $F=\left[F_{1}, F_{2}, F_{3}, \cdots, F_{i}, \cdots, F_{n}\right]$ and the subspace dimension $k$. Then the training set $F$ is centered.
		\STATE Initialization: Compute standard 2DPCA and obtain $W_{0}$. Set $W=W_{0}$.
		\STATE Calculate the Cauchy weight:
		
		Compute residue $s_{i}=\left\|F_{i}-W W^{T} F_{i}\right\|_{F}$.
		
		Compute $c=\operatorname{median}\left(s_{i}\right)$.
		
		Compute $\omega_{i}^{(C)}=\left(1+\left\|\mathrm{F}_{i}-W W^{T} F_{i}\right\|_{F}^{2} / c^{2}\right)^{-1}$
		
		\STATE Calculate the covariance matrix: $C_{r}=\sum_{i=1}^{n} \omega_{i} F_{i} F_{i}^{T}$
		\STATE Update $W$:
		
		$W^{\left(t+\frac{1}{2}\right)}=C_{r}\left(W^{(t)}\right) W^{(t)}$
		
		$W^{(t+1)}=$ orthoronalize $\left(W^{\left(t+\frac{1}{2}\right)}\right)$
		\STATE Convergence check:
		
		If $W^{t+1} \neq W^{t}$, go to Step 4.
		
		Else go to Step 6.
		\STATE Calculate the uncentered data: $\forall i, F_{i}=F_{i}+\bar{F}$
		\STATE Projection: $V=W^{T} X$.
		\ENSURE $W$ and $V$.
	\end{algorithmic} 
\end{algorithm}

\subsection{2DL$_1$-PCA}
\label{sec:3.2}
Compared to L$_1$-PCA, in the two dimensional case we want to find a column vector to solve the following problem
\begin{equation}\label{equ:14}
w^{*}=\max _{w}\left\|w^{T} F\right\|_{L_{1}}=\max _{w} \sum_{i=1}^{n}\left\|w^{T} F_{i}\right\|_{L_{1}}, \text { subject to }\|w\|_{2}=1
\end{equation}

In fact, $w^{T}F$ is a row vector. The number of maximum absolute value in a vector contributes most to its L$_1$ norm. Assume that the column index of the maximum absolute value in $w^{T}F$ is $i$, we can calculate $w^{*}$ by the $i$th column of $F_{i}$. The 2DL$_1$-PCA algorithm is given in Algorithm~\ref{alg:4}.
\begin{algorithm} 
	\caption{The 2DL$_1$-PCA algorithm} 
	\label{alg:4} 
	\begin{algorithmic}[1]
		\REQUIRE The training set $F=\left[F_{1}, F_{2}, F_{3}, \cdots, F_{i}, \cdots, F_{n}\right]$.
		\STATE Initialization: Pick any $w_0$. set $w(0)=w(0) /\|w(0)\|_{2}$, and $t=0$.
		\STATE Polarity check: For all $i \in\{1, \cdots, n\}, v=w^{T}(t) F_{i},[mv~mi]=\max (abs(v)), q_{i}(t)=mi$, if $v(mi)<0, p_{i}(t)=-1, $else $p_{i}(t)=1$.
		\STATE Flipping and maximization: Set $t=t+1$, and 
		$w(t)=\sum_{i=1}^{n} p_{i}(t-1) F_{i}\left(:, q_{i}(t-1)\right)$, Set $w(t)=w(t) /\|w(t)\|_{2}$.
		\STATE Convergence check:
		
		If $w(t) \neq w(t-1)$, go to step 3.
		
		Else if $i$ exists such that $w^{T}(t)F_{i}\left(:, q_{i}\left(t\right)\right)=0$, set $w(t)=(w(t)+\Delta w(t)) /\|w(t)+\Delta w\|_{2}$, and go to step 3. Here $\Delta w$ is a small nonzero random vector.
		
		Otherwise, set $w^{*}=w(t)$ and stop.
		\ENSURE The projection vector $w$.
	\end{algorithmic} 
\end{algorithm}

Then we can obtain a $k$ dimensional projection matrix from the following algorithm.

$w_{0}=\mathbf{0},\left\{F_{i}^{0}=F_{i}\right\}_{i=1}^{n}$.

For $j=1$ to $k$

$\quad \forall i \in\{1,2,3 \cdots, n\}, F_{i}^{j}=F_{i}^{j-1}-w_{j-1}\left(w_{j-1}^{T} F_{i}^{j-1}\right)$.

$\quad$ Apply the L$_1$-PCA procedure to $F^{j}=\left[F_{1}^{j}, \cdots, F_{n}^{j}\right]$ to find $w_{j}$.

End

\section{Experimental results and analysis}
\label{sec:4}
Three databases: ORL, Yale and XM2VTS are used to test methods mentioned above. The recognition accuracy and running time of extracting features are recorded.

The ORL database contains face images from 40 different people and each person has 10 images, the resolution of which is 92$\times$112. Variation of expression (smile or not) and face details (wear a glass or not) are contained in the ORL database images. In the following experiments, 5 images are selected as the training samples and the rest are selected as the test samples.

The Yale database is provided by Yale University. This database contains face images from 15 different people and each has 11 images. The resolution of Yale database images is 160$\times$121. In the following experiments, 6 images are selected as the training samples and the rest are selected as the test samples.

The XM2VTS\cite{messer1999xm2vtsdb} database offers synchronized video and speech data as well as image sequences allowing multiple view of the face. It contains frontal face images taken of 295 subjects at one month intervals taken over a period of few months. The resolution of XM2VTS is 55$\times$51. In the following experiments, 4 images are selected as the training samples and the rest are selected as the test samples.

\subsection{R$_1$-PCA and 2DR$_1$-PCA}
\label{sec:4.1}
The experimental results of R$_1$-PCA and 2DR$_1$-PCA are shown in Table 1, and the number of iterations of R$_1$-PCA and 2DR$_1$-PCA is 120.

\begin{table}[h]
\label{tab:tab-1}
\centering
\caption{Experimental results of R$_1$-PCA and 2DR$_1$-PCA}
\begin{tabular}{cccc}
\hline
            & ORL    & Yale   & XM2VTS   \\ \hline
\multicolumn{4}{l}{Recognition accuracy} \\ \hline
PCA         &    0.90    &  0.77      &   0.71       \\
R$_1$-PCA      &   0.88     &   0.77     &   0.71       \\
2DR$_1$-PCA    &    0.90    &    0.80    &     0.78     \\ \hline
\multicolumn{4}{l}{Running time}         \\ \hline
PCA         &   1.37     &    0.36    &    17.27      \\
R$_1$-PCA      &  914.21      &   411.06     &   1409.30       \\
2DR$_1$-PCA    &   403.90     &   372.76     &   619.78       \\ \hline
\end{tabular}
\end{table}

The initial projection matrix $W^{0}$ is obtained by PCA (2DPCA) at the beginning of R$_1$-PCA (2DR$_1$-PCA). The final projection matrix $W$ is obtained by an iterative method starting with $W^{0}$. As a result of the iteration, the computational complexity is high. Meanwhile, they have nearly the same recognition accuracy.

\begin{figure}[htbp]
	\centering
	\includegraphics[trim={0mm 0mm 0mm 0mm},clip, width = .8\textwidth]{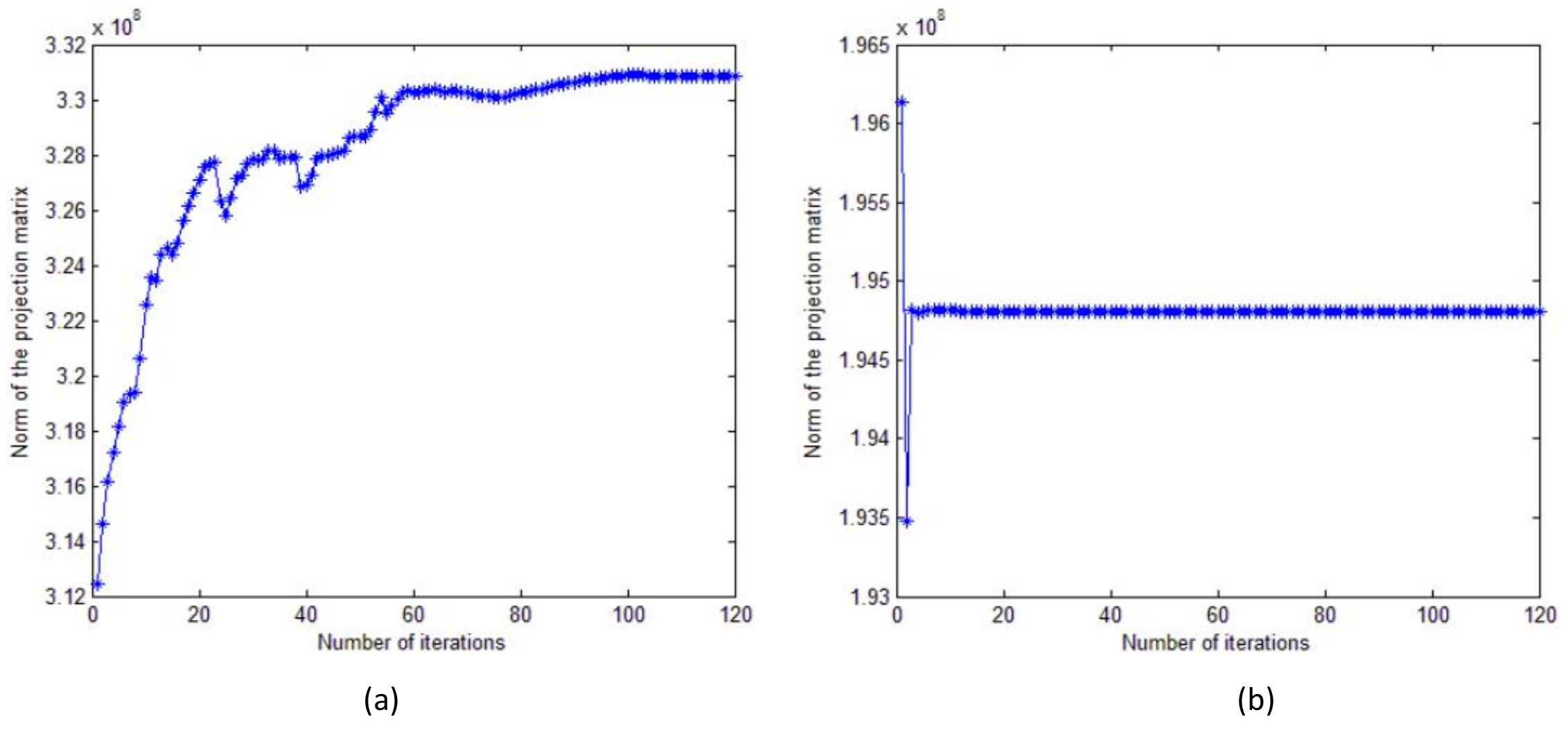}
	\caption{The convergence illustration of iterating 120 times on the ORL database. (a) R$_1$-PCA. (b) 2DR$_1$-PCA.}
	\label{fig:orl}
\end{figure}

In the experiment of R$_1$-PCA algorithm tested on the ORL database, the convergence process is shown in Fig.~\ref{fig:orl} (a), in which the $y$-coordinate denotes the norm of projection matrix and the $x$-coordinate denotes the number of iterations. The norm of a projection matrix is used to observe its convergent process. After iterating at least 100 times the projection matrix $W$ converges. As a comparison, 2DR$_1$-PCA just needs less than 30 iteration to obtain a convergent projection matrix, which is shown in Fig.~\ref{fig:orl} (b). Image matrices used in 2DR$_1$-PCA leads to a faster convergence.

\begin{figure}[htbp]
	\centering
	\includegraphics[trim={0mm 0mm 0mm 0mm},clip, width = .8\textwidth]{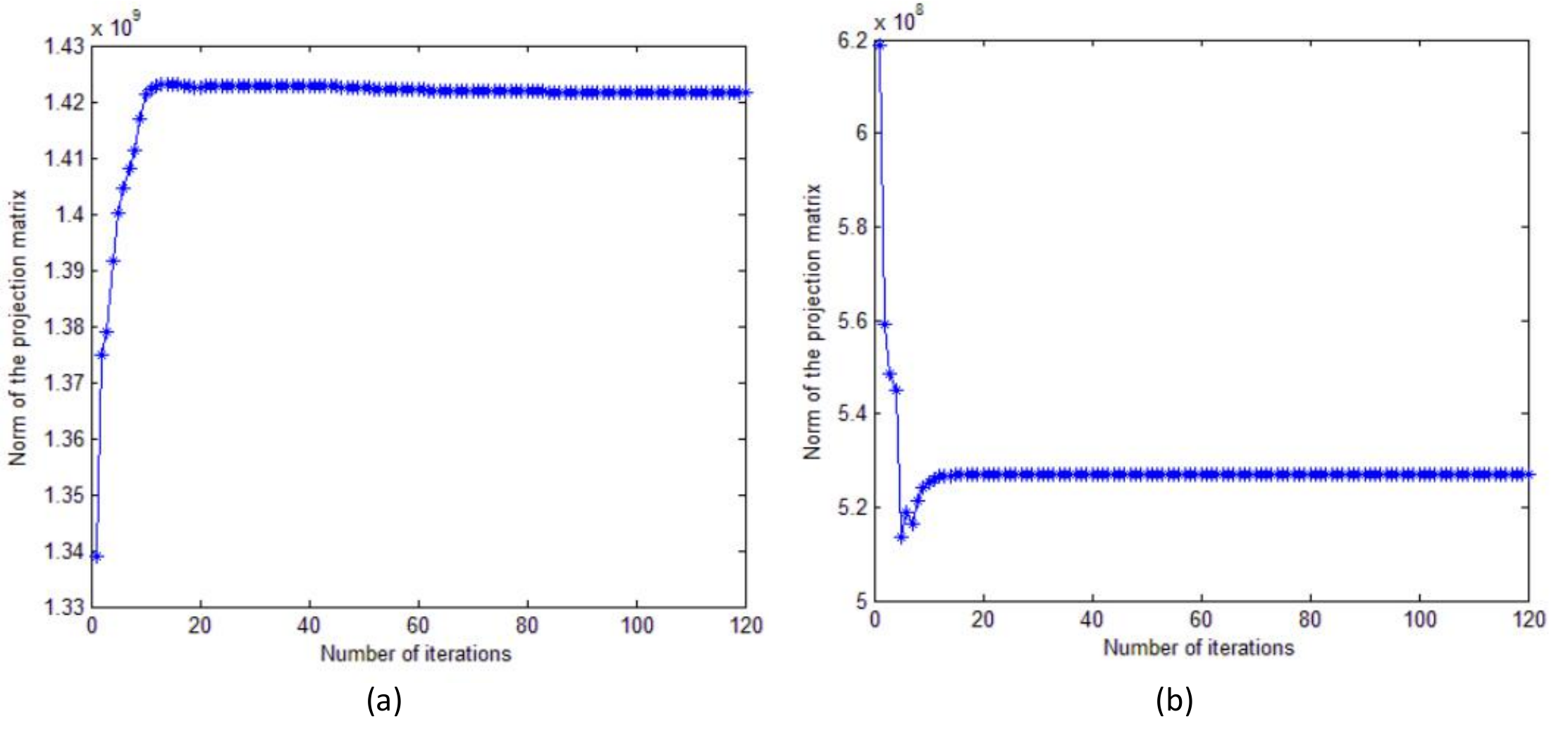}
	\caption{The convergence illustration of iterating 120 times on the Yale database. (a) R$_1$-PCA. (b) 2DR$_1$-PCA.}
	\label{fig:yale}
\end{figure}

\begin{figure}[htbp]
	\centering
	\includegraphics[trim={0mm 0mm 0mm 0mm},clip, width = .8\textwidth]{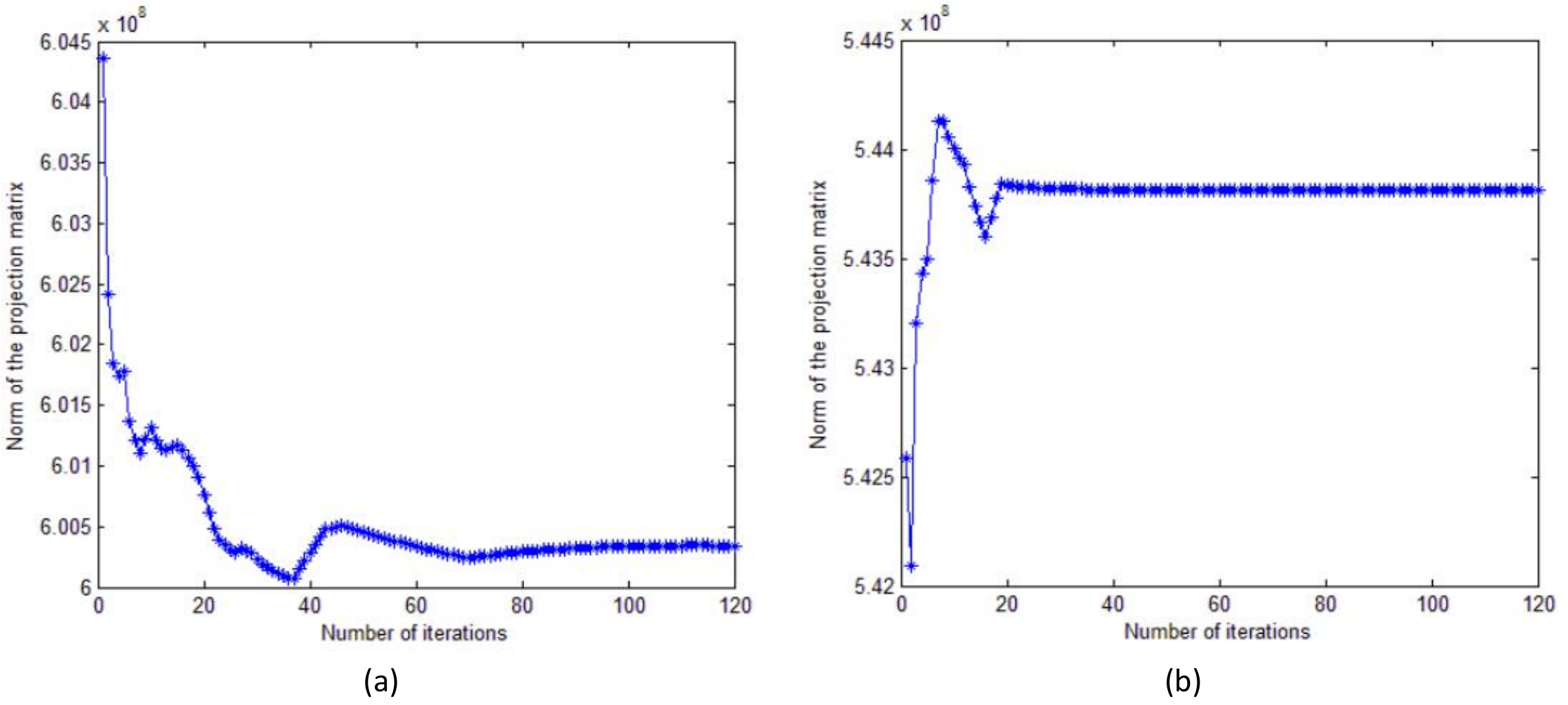}
	\caption{The convergence illustration of iterating 120 times on the XM2VTS database. (a) R$_1$-PCA. (b) 2DR$_1$-PCA.}
	\label{fig:xm2}
\end{figure}

The convergence illustration tested on the Yale database is shown in Fig.~\ref{fig:yale}. The convergent speed of R$_1$-PCA is similar to that of 2DR$_1$-PCA. In the experiment tested on the XM2VTS database, the convergent speed of 2DR$_1$-PCA is much faster than that of R$_1$-PCA shown in Fig.~\ref{fig:xm2}. In other words, the efficiency of 2DR$_1$-PCA is higher than that of R$_1$-PCA.

\subsection{L$_1$-PCA and 2DL$_1$-PCA}
\label{sec:4.2}
The experimental results of L$_1$-PCA and 2DL$_1$-PCA are shown in Table 2.

\begin{table}[h]
\label{tab:tab-1}
\centering
\caption{Experimental results of L$_1$-PCA and 2DL$_1$-PCA}
\begin{tabular}{cccc}
\hline
            & ORL    & Yale   & XM2VTS   \\ \hline
\multicolumn{4}{l}{Recognition accuracy} \\ \hline
PCA         &    0.90    &  0.77      &   0.71       \\
L$_1$-PCA      &   0.90     &   0.76     &   0.71       \\
2DL$_1$-PCA    &    0.91    &    0.80    &     0.76     \\ \hline
\multicolumn{4}{l}{Running time}         \\ \hline
PCA         &   1.37     &    0.36    &    17.27      \\
L$_1$-PCA      &  15.96      &   5.15     &   83.52      \\
2DL$_1$-PCA    &   3.52     &   3.62    &   40.43      \\ \hline
\end{tabular}
\end{table}

From Table 2 we can see that the performance of 2DL$_1$-PCA is better than that of L$_1$-PCA and PCA. In 2DL$_1$-PCA, image matrices are used directly for feature extraction. Features extracted by 2DL$_1$-PCA is less than features extracted by L$_1$-PCA.

\begin{figure}[htbp]
	\centering
	\includegraphics[trim={0mm 0mm 0mm 0mm},clip, width = .8\textwidth]{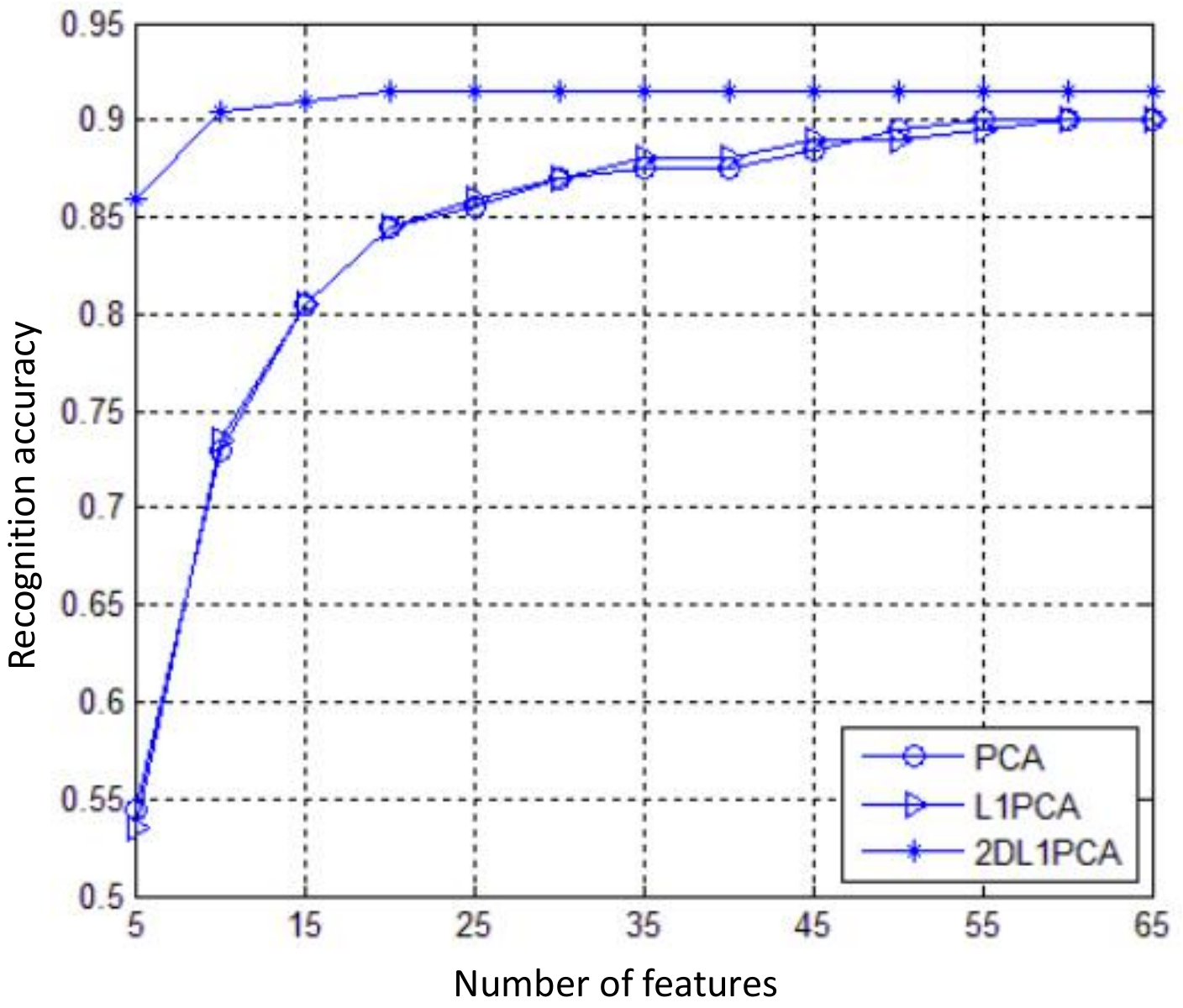}
	\caption{Recognition accuracy versus different number of features on the ORL database.}
	\label{fig:accuracy}
\end{figure}

We implement another experiment on the ORL database. Different number of features is extracted by PCA, L$_1$-PCA and 2DL$_1$-PCA, respectively. Then these features are used for face recognition. The experimental result is shown in Fig.~\ref{fig:accuracy}, from which we can see that less features extracted by 2DL$_1$-PCA achieves a higher recognition accuracy.

\section{Conclusions}
\label{sec:5}
In this paper we proposed 2DR$_1$-PCA and 2DL$_1$-PCA for face recognition. We extend R$_1$-PCA and L$_1$-PCA to their 2-D case so that image matrices could be directly used for feature extraction. Compared to the  L$_2$ norm based methods, these L$_1$ norm based methods are less sensitive to outliers. We analyze the performance of 2DR$_1$-PCA and 2DL$_1$-PCA against R$_1$-PCA and L$_1$-PCA algorithms based on experiments. The experimental results show that the performance of 2DR$_1$-PCA and 2DL$_1$-PCA is better than that of R$_1$-PCA and L$_1$-PCA, respectively.

\acknowledgments
This work was partially supported by the National Natural Science Foundation of China (Grant No.61672265 and U1836218) and the 111 Project of Ministry of Education of China (Grant No. B12018). 


\bibliography{report}   
\bibliographystyle{spiejour}   


\listoffigures
\listoftables

\end{spacing}
\end{document}